# Hybrid Transformer and Spatial-Temporal Self-Supervised Learning for Long-term Traffic Prediction

**Wang Zhu**
School of Optical-Electrical and Computer Engineering
University of Shanghai for Science and Technology, Shanghai, China, 200093
Email: 223330859@st.usst.edu.cn

**Doudou Zhang**
Principal Data Scientist
Capital One
Mclean, VA, USA, 22102
Email: Doudouzhang181@gmail.com

**Baichao Long**
School of Optical-Electrical and Computer Engineering
University of Shanghai for Science and Technology, Shanghai, China, 200093
Email: 212190376@st.usst.edu.cn

**Jianli Xiao (Corresponding Author)**
School of Optical-Electrical and Computer Engineering
University of Shanghai for Science and Technology, Shanghai, China, 200093
Email: audyxiao@sjtu.edu.cn

Word Count: 5589 words + 3table (250 words per table) = 6339 words

## ABSTRACT

Long-term traffic prediction has always been a challenging task due to its dynamic temporal dependencies and complex spatial dependencies. In this paper, we propose a model that combines hybrid Transformer and spatio-temporal self-supervised learning. The model enhances its robustness by applying adaptive data augmentation techniques at the sequence-level and graph-level of the traffic data. It utilizes Transformer to overcome the limitations of recurrent neural networks in capturing long-term sequences, and employs Chebyshev polynomial graph convolution to capture complex spatial dependencies. Furthermore, considering the impact of spatio-temporal heterogeneity on traffic speed, we design two self-supervised learning tasks to model the temporal and spatial heterogeneity, thereby improving the accuracy and generalization ability of the model. Experimental evaluations are conducted on two real-world datasets, PeMS04 and PeMS08, and the results are visualized and analyzed, demonstrating the superior performance of the proposed model.

**Keywords:** Transformer, Self-supervised Learning, Graph Convolution Network, Traffic Prediction

**INTRODUCTION**

In recent decades, there has been significant development in transportation infrastructure. However, due to the increasing urban population and the growing number of vehicles in cities, traffic problems continue to escalate, leading to a rise in congestion-related issues (*1*). Traffic congestion not only exacerbates air pollution and fuel consumption but also increases the occurrence of accidents and adds to the cost of travel. Traffic prediction plays a crucial role in intelligent transportation systems (*2*), combining technologies such as big data analytics and intelligent algorithms to extract useful features from multiple data sources. This optimization helps in improving traffic signal control, route selection, and enhancing road utilization, thereby alleviating congestion issues (*3*). Traffic speed, which provides real-time information on traffic congestion, is one of the key indicators in traffic prediction. Traffic speed prediction plays a significant role in intelligent transportation systems (*4*).

Currently, most models for traffic prediction are based on deep learning techniques. However, many of these studies are limited to short-term traffic prediction, with most forecasting time horizons ranging within half an hour, lacking medium to long-term prediction capabilities. In traffic prediction, the dynamic nature of traffic conditions and external influencing factors pose a greater challenge for medium to long-term prediction. In daily travel, timely and accurate medium to long-term traffic prediction can provide travelers with more data support for making informed decisions. As shown in **Figure 1**, while Route B does not experience congestion in the short term, congestion occurs when vehicles reach the next observation point. On the other hand, Route A provides a longer prediction time, allowing travelers to effectively avoid congested sections during their journey. Therefore, providing medium to long-term predictions is beneficial for making informed travel decisions and improving travel efficiency (*5*).

Long-term traffic prediction has always been a challenging task due to its dynamic temporal dependencies, complex spatial dependencies, and sensitivity to errors. In recent years, with the rapid development of neural networks, many studies have used recurrent neural networks (RNNs) (*6*) and their variants such as long short-term memory (LSTM) (*7*) to model the temporal scale and extract underlying temporal correlations. Concurrently, convolutional neural networks (CNNs) (*8*) and graph neural networks (GNNs) (*9*) have been employed to model spatial scale and extract spatial correlations for Euclidean and non-Euclidean data, respectively. Long Sequence Time Series Forecasting (LSTF) requires models to effectively capture the long-term dependency coupling in data (*10*). However, these models still face challenges in capturing long-term dependencies and have limitations in parallel computing capabilities. In contrast, the Transformer model, with its self-attention mechanism, allows for parallel computation of correlations between nodes, thereby improving the efficiency of training and prediction. Compared to RNN models, Transformer has shown better performance in capturing long-range dependencies (*11*). Due to the spatio-temporal heterogeneity (*12*) in traffic patterns, there is often a significant deviation in traffic speed between different functional regions at the same time and within the same region at different time periods. However, existing models mostly rely on prior knowledge of traffic spatio-temporal patterns, making predefined annotations of external factors and grid density, which limits the generalization ability of the models (*13*).

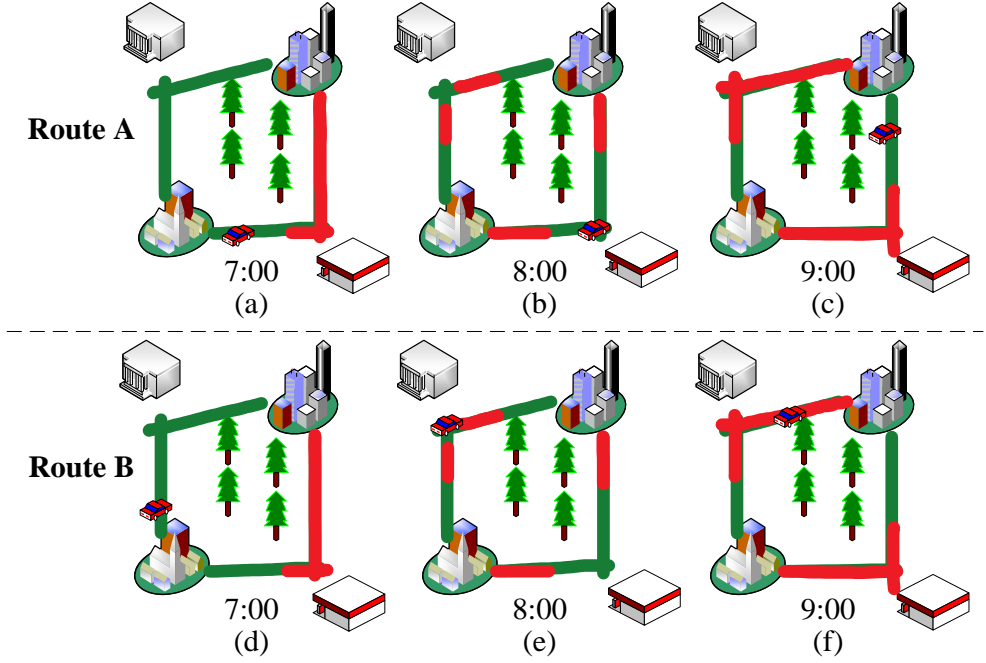

**Figure 1 Routes planning. (a), (b), and (c) represent the congestion situation of route A at three different time periods; (d), (e), and (f) represent the congestion situation of route B at three different time periods**

To address the aforementioned challenges and improve the model's generalization ability while making real-time and accurate long-term traffic predictions, this paper proposes a novel traffic prediction model that hybrid Transformer and spatio-temporal self-supervised learning(T-ST-SSL). The main contributions of this paper are as follows:

1) We design a spatio-temporal block that incorporates both graph convolution and Transformer, allowing for the extraction and aggregation of spatial dependencies and long-term temporal dependencies underlying traffic patterns.

2) We Introduce a spatio-temporal self-supervised learning framework that effectively mitigates noise and perturbations using data augmentation techniques. Furthermore, the framework enhances the model's ability to handle spatio-temporal heterogeneity through self-supervised learning tasks.

3) We conducted extensive experiments on two real-world datasets. When compared with seven baseline models, our T-ST-SSL model consistently exhibited the best predictive performance across four evaluation metrics.

The remaining structure of this paper is organized as follows: In Section 2, we review some typical spatio-temporal prediction models and introduce self-supervised learning. Then, in Section 3, we define the problem of traffic prediction and provide a preliminary introduction to the related parameters. Subsequently, in Section 4, we provide a detailed description of the individual modules of the T-ST-SSL model. Finally, in Section 5, we conduct comparative experiments between our proposed model and seven baseline models on two real-world datasets. We also perform ablation experiments to validate the effectiveness of each module in our model.

To visually demonstrate the predictive performance of our proposed model, we visualized the prediction results.

## RELATED WORKS
### Deep Learning Models

Deep learning models have made significant progress in traffic prediction and have become a hot topic in both research and practical applications. These models are capable of effectively extracting hidden features from traffic data by modeling complex spatio-temporal relationships, leading to accurate predictions. Addressing the issue of extracting spatial dependencies mainly from fixed graph structures, Wu et al. (*14*) introduced an adaptive matrix that captures spatial dependencies accurately through node embedding learning. Zhao et al. (*15*) combined graph convolutional networks (GCN) and gated recurrent units (GRU) to model the topology of road networks and the dynamic changes in traffic data, allowing for simultaneous capture of spatio-temporal dependencies. In recent years, with the introduction of Transformer (*16*), they have been widely applied in long sequence time prediction due to their parallel computation capability and ability to capture global information. Jiang et al. (*17*) designed a spatial self-attention module to capture dynamic spatial dependencies and fuse local and global information using different masking methods, enabling simultaneous capture of short-range and long-range spatial dependencies. Xu et al. (*18*) jointly modeled multiple factors' relationships through a multi-head attention mechanism and different spatial dependency patterns, and employs a temporal Transformer to model remote bidirectional time dependencies, enabling fast and scalable training on long-term spatio-temporal dependencies.

### Self-Supervised Learning Models

The goal of self-supervised learning is to improve representation quality by leveraging the underlying information in the data through various pretext tasks (*19*). Compared to traditional methods that require a large amount of manually annotated labeled data, self-supervised learning is an effective approach for addressing data scarcity issues as it does not rely on a significant amount of labeled data during the training of neural network models. In situations with the same quantity of labeled data, self-supervised learning can assist in training more accurate models (*20*). Self-supervised learning has been widely applied and achieved significant success in computer vision, natural language processing, and other fields. Wickstrøm et al. (*21*) achieved transfer learning for time series by introducing the concept of label smoothing, which improved the performance of model classification. Bielak et al. (*22*) proposed a novel framework for self-supervised graph representation learning, which optimizes representation vectors by computing the cross-correlation matrix of embeddings from two distorted views of an individual graph, achieving remarkable success in node classification tasks. In this paper, through adaptive augmentation of spatio-temporal graph data and the introduction of two pretext tasks, we train the model on more granular labels, learning the spatio-temporal heterogeneity of traffic speed data, thereby enhancing model performance.

## PRELIMINARIES
### Definition

We define the traffic road network graph as $G = (V, E, A)$, where $V = \{v_1, \cdots, v_n\}$

represents the set of spatial regions (nodes) in the road network, with a size of $|V| = N$. $E$ denotes the set of edges between spatial regions, and $A \in R^{N \times N}$ is the adjacency matrix of the roads. The weights are determined by calculating the Euclidean distance between regions, indicating the connectivity between them. We use $x_i \in R^{N \times d}$ to represent the traffic speed at time $t$ for each node, where $d$ is the number of features for traffic information. In this paper, we set $d$ to 1, using traffic speed as the feature for prediction. We represent the feature matrix for $T$ time slices of node speeds as $X = [x_{(t-T+1)}, ..., x_t] \in R^{T \times N \times d}$.

**Problem Statement**

Traffic prediction is a typical spatiotemporal forecasting problem aimed at predicting future traffic information based on historical observations. Formally, given the historical traffic information $X = [x_{(t-T+1)}, ..., x_t] \in R^{T \times N \times d}$ for $T$ time slices and the road network graph $G$, the goal is to learn a mapping function $F$ that can predict the traffic information at time slot $t_p$.

$$\hat{y}_{t_p} = F[(x_{(t-T+1)}, ..., x_t); G] \qquad (1)$$

**METHODS**

In this section, we will provide a detailed introduction to the hybrid Transformer and spatial-temporal self-supervised learning network model. Specifically, we will outline the overall framework of the model and then elaborate on its components, including the time Transformer block, spatial graph convolution block, adaptive graph augmentation module, spatial-temporal heterogeneity self-supervised learning module, and prediction module.

**The Overall Framework**

The overall framework of the hybrid Transformer and spatial-temporal self-supervised learning network model proposed in this paper is shown in **Figure 2**. The network primarily consists of stacked spatial-temporal blocks and spatial-temporal heterogeneity measurement blocks. Firstly, the historical traffic speed information is augmented through the adaptive graph augmentation module to obtain a more comprehensive spatial-temporal information. Then, the original data and augmented data are separately fed into the stacked spatial-temporal blocks. Each spatial-temporal block consists of a time Transformer module and a spatial graph convolution module, which capture the spatial-temporal dependencies in the time and space dimensions and jointly learn latent spatial-temporal features. By stacking multiple spatiotemporal blocks, the depth of the network model is increased to acquire more complex and accurate spatiotemporal features. Finally, the prediction module aggregates and predicts the learned spatial-temporal features, while the spatial-temporal heterogeneity self-supervised learning module is utilized to learn spatial-temporal heterogeneity. The prediction loss and the loss from the self-supervised tasks are aggregated, and the model is trained using the backpropagation algorithm.

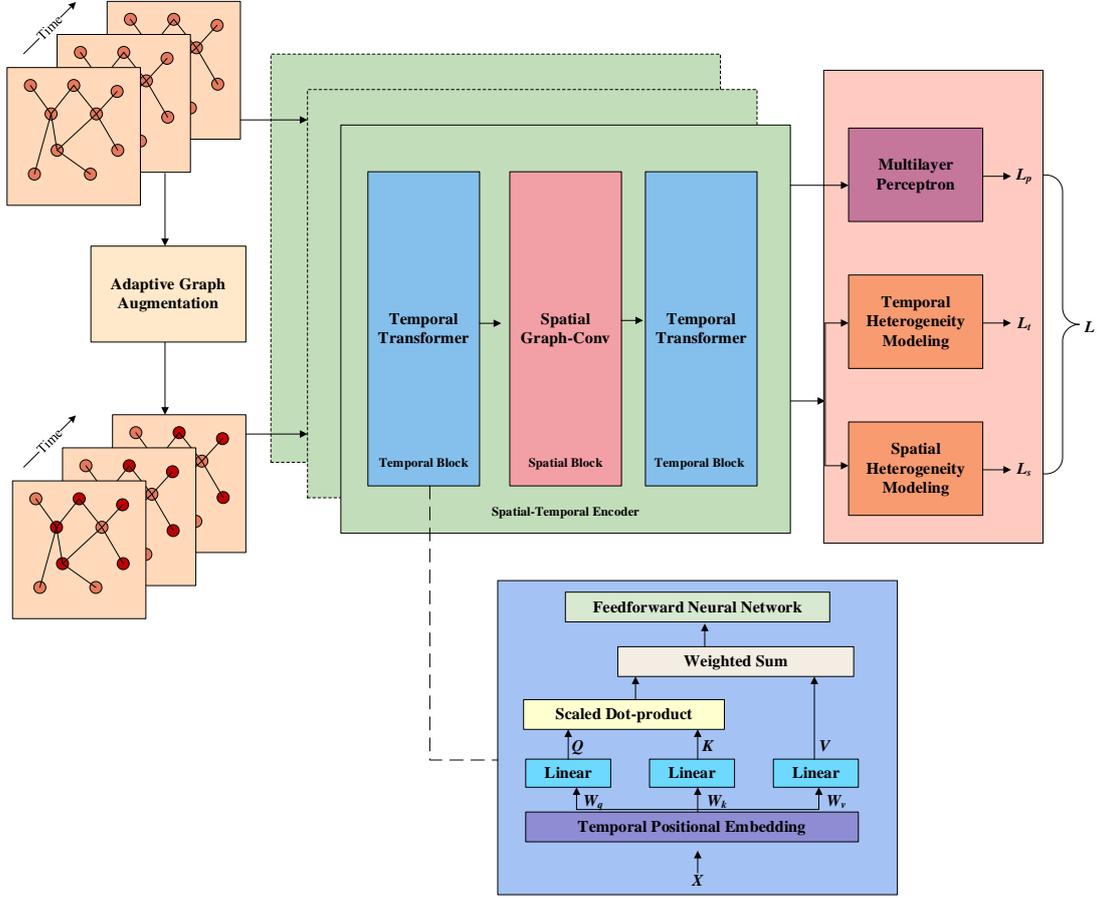

**Figure 2 The overall framework of the proposed model**

## Spatial-Temporal Encoder

Traffic prediction is a typical spatio-temporal forecasting problem, where traffic speed not only changes dynamically over time but is also influenced by factors such as the environment and road network conditions. To better capture the hidden spatio-temporal dependencies in traffic information and achieve real-time and accurate long-term traffic speed prediction, we propose a spatial-temporal encoder that combines Transformer and graph convolution. As shown in **Figure 2**, we stack multiple spatial-temporal encoders to extract deep spatio-temporal dependencies. $X^K \in R^{T \times N \times d}$ and the traffic graph $G$ are the inputs to the $K$ th spatial-temporal encoder. The output of the $K$ th spatial-temporal encoder is $X^{K+1} \in R^{T \times N \times d}$.

$$X^{K+1} = f(X^K; G) \qquad (2)$$

By stacking multiple spatial-temporal encoders and incorporating gated linear units (GLUs), we control the information flow and selection to generate the final embedding matrix $X^{ST}$ of the ST encoder.

$$X^{ST} = (Conv(X^{K+1}) + X^{K+1}) * sigmoid(Conv(X^{K+1})) \qquad (3)$$

Where $X^{ST} \in R^{T \times N \times d}$, $T$ is the length of the time steps and $d$ represents the dimensionality of the embeddings. The $n$ th row $x_n^{ST} \in R^{T \times d}$ represents the final embedding of region $v_n$.

**Temporal Block**

As shown in the bottom of **Figure 2**, we propose a Transformer-based temporal module. Compared to RNNs and its variants, this module utilizes self-attention mechanism to model relationships between different positions, allowing for better capture of long-term temporal dependencies. Moreover, it enables parallel computation, making it more effective for processing long sequences of data. Specifically, we project the input features $X \in R^{T \times N \times d}$ from the traffic graph information into a high-dimensional subspace through learned linear mappings. Then, we dynamically compute the temporal dependencies using the self-attention mechanism. The query vector $Q^T$, key vector $K^T$, and value vector $V^T$ can be computed as follows:

$$\begin{cases} Q^T = XW_Q^T \\ K^T = XW_K^T \\ V^T = XW_V^T \end{cases} \qquad (4)$$

Where $W_Q^T$, $W_K^T$, $W_K^T$ are learnable parameters representing the learned linear mappings. Then, we use the scaled dot-product function to convert the results into attention scores, which represent the temporal dependency.

$$S^T = soft \max(Q^T (K^T)^{\mathrm{T}} / \sqrt{d_k^T}) \qquad (5)$$

Then, we aggregate the calculated attention scores with $V^T$ and feed them into a feedforward neural network to extract relationships between deeper layers of features.

$$X^T = \mathrm{Re}\, Lu(\mathrm{Re}\, Lu(M^T W_0^T) W_1^T) W_2^T \qquad (6)$$

Where $M^T = S^T V^T + X^T$, residual connections are used to propagate information between different layers and mitigate feature loss. $X^T$ represents the region embedding matrix, where the $n$ th row $x_n^T$ corresponds to the embedding of region $v_n$.

**Spatial Block**

Graph convolution is a generalization of classical convolution to the graph domain, used to extract topological information in road networks. In this paper, we propose a Spatial Convolution (SC) block that employs Chebyshev graph convolution to aggregate node information from roads and capture spatial dependencies. Firstly, we compute the adjacency matrix $A \in R^{N \times N}$ based on the Euclidean distance between nodes in the topological graph. Then, we calculate the degree matrix $D$ and the Laplacian matrix $\Gamma = 2L / \lambda_{\max} - I_n$ scaled by Chebyshev polynomials, based on the adjacency matrix. Where $L = I_n - D^{-1/2}AD^{-1/2}$ is the normalized Laplacian matrix, $I_n$ is the unit matrix, $D_{ii} = \sum_i A_{ij}$, $\lambda_{\max}$ represents the maximum eigenvalue of $L$. By setting the output $X^T$ of the temporal block as the input, $T_k$ represents the $k$ th order Chebyshev polynomial. The $K$ th order Chebyshev polynomial graph convolution effectively extracts spatial topological information from the road network.

$$X^S = \sum_{i=1}^{d} \sum_{k=0}^{K} \theta T_k(\Gamma) X_{:,i}^T \qquad (7)$$

Where $\theta$ is the learnable parameter，$X_{:,i}^T$ represents the $i$ th channel of the node features.

**Adaptive Graph Augmentation Module**

To better extract spatio-temporal dependencies in traffic information and aid the model in learning invariance and variability in the data, thereby improving its generalization ability and accuracy, we perform sequence-level and topological structure-level data augmentation on the input sequence $X \in R^{T \times N \times d}$ and the topological graph $G$.

Firstly, for each region in the road network graph, we need to perform heterogeneity measurement. For region $v_n$, we utilize the embedded sequence $x_n^T \in R^{T \times d}$ from the $n$ th row of the output $X^T \in R^{T \times N \times d}$ of the time block to generate the overall traffic variation embedding:

$$r_n = \sum_{\tau=t-T}^{t} u_{\tau,n} \cdot x_{\tau,n}^T, \quad where \ u_{\tau,n} = (x_{\tau,n}^T)^T \cdot w_0 \qquad (8)$$

$r_n$ represents the aggregated representation of region $v_n$ at different time slices, which is used

to further calculate the heterogeneity level between regions. $u_{\tau,n}$ is the aggregation weight of the embedding sequence $x_n^T$, reflecting the correlation between local temporal traffic speed states and overall traffic speed state $w_0$ is the learnable parameter $\tau$ is the index of the time step range $(t-T, t)$. To better capture the differences in traffic speed and temporal variations between different regions, we calculate the heterogeneity level between regions using the **Equation 9**:

$$\eta_{m,n} = \frac{r_m^{\mathrm{T}} r_n}{\|r_m\| \|r_n\|} \tag{9}$$

Based on the magnitude of the score in $\eta_{m,n}$, we can visually distinguish the heterogeneity levels between different regions. A higher score indicates a stronger correlation in speed between region $v_m$ and $v_n$, indicating a lower level of heterogeneity.

*Traffic Sequence-level Augmentation*

In order to better capture the correlation between region specific traffic speed sequences and the overall traffic speed patterns, we employ a mask probability $\mu_{\tau,n} \sim Bern(1 - u_{\tau,n})$ that follows a Bernoulli distribution to manipulate the traffic sequence in region $v_n$ at the $\tau$ th time step. Specifically, a larger value of $\mu_{\tau,n}$ indicates a weaker correlation between the traffic speed $x_{\tau,n}$ at the $\tau$ th time step and the overall traffic speed states of that region, making it more likely to be masked and highlighting the data with higher correlation. The augmented data obtained after this manipulation is denoted as $\tilde{X} \in R^{T \times N \times d}$ .

*Graph Topological Structure-level Augmentation*

The traffic patterns between different functional areas have a significant impact on traffic speed prediction. In the same time period, areas with similar functions often exhibit similar traffic patterns. A single adjacency matrix is insufficient to capture spatial correlations in traffic information. Therefore, we propose graph topological structure-level augmentation. For two adjacent areas, $v_m$ and $v_n$, we eliminate connections between regions with low correlation in traffic patterns. The mask probability follows a Bernoulli distribution $\mu_{m,n} \sim Bern(1 - \eta_{m,n})$. A smaller measure of heterogeneity $\eta_{m,n}$ indicates a lower inter-region correlation, leading to a higher likelihood of masking their connections. For two non-adjacent regions, we enhance connections between regions with high correlation. The mask probability follows a Bernoulli distribution $\mu_{m,n} \sim Bern(\eta_{m,n})$. A larger measure of heterogeneity $\eta_{m,n}$ indicates a higher inter-region correlation, leading to the addition of connections between them. This approach not only eliminates low correlation interference but also captures spatial features with long-term dependencies. The augmented graph data obtained through this process is denoted as $\tilde{G} = (V, \tilde{E}, \tilde{A})$.

**Spatial Heterogeneity Self-Supervised Learning Module**

The traffic patterns between different functional areas are often distinct. To accurately capture long-term spatial correlations, it is crucial to model spatial heterogeneity accurately. Therefore, we propose a self-supervised learning task to improve the quality of representations. By encoding augmented traffic graph data and predicting the clustering assignment of different functional areas, we effectively preserve spatial heterogeneity. Additionally, we design an auxiliary learning task to compute the prediction allocation score, enabling each region to reflect its spatial heterogeneity compared to the overall region. First, we define $K$ clusters to represent different functional areas, with the embedding representation of the $K$ clusters denoted as $\{q_1, \cdots, q_K\}$. Next, we compute the correlation score $\tilde{p}_{n,k} = q_k^\top \tilde{x}_n^{ST}$ and the prediction allocation score $\hat{p}_{n,k} = q_k^\top x_n^{ST}$ between the embedding representation of region $v_n$ and the embedding representation of the $k$ clusters. Here $\tilde{x}_n^{ST}$ represents the embedding representation of region $v_n$ encoded from augmented traffic graph data, and $x_n^{ST}$ represents the embedding representation of region $v_n$ encoded from the original traffic graph data. Finally, the optimization function of the self-supervised learning task is defined as follows:

$$l(x_n^{ST}, \tilde{p}_n) = -\sum_k \tilde{p}_{n,k} \log \frac{\exp(\hat{p}_{n,k} / \theta)}{\sum_j \exp(\hat{p}_{n,j} / \theta)} \qquad (10)$$

Where $\theta$ is the temperature parameter used to adjust the smoothness of the softmax function's output. The self-supervised task for capturing the spatial heterogeneity of the overall region is defined as follows:

$$L_s = \sum_{n=1}^N l(x_n^{ST}, \tilde{p}_n) \qquad (11)$$

**Temporal Heterogeneity Self-Supervised Learning Module**

The traffic patterns of the same region are typically different across different time periods. For example, there is heavy traffic during peak hours and less traffic during nighttime at the same region. Therefore, it is necessary to model not only spatial heterogeneity but also temporal heterogeneity. We consider positive pairs as the region embeddings and overall embeddings from the same time step, and negative pairs as the region embeddings and overall embeddings from different time steps. Positive pairs are used to capture the consistency of overall traffic trends, while negative pairs are used to model the temporal heterogeneity across different time steps. We start by integrating the region embeddings obtained from both the original traffic data and the augmented traffic data.

$$r_{t,n} = w_1 \odot x_{t,n}^{ST} + w_2 \odot \tilde{x}_{t,n}^{ST} \qquad (12)$$

Where $w_1$, $w_2$ are learnable parameters, and $\odot$ represents element-wise multiplication. Then, by aggregating the embeddings of each region, we generate the overall embedding representation for time step $t$.

$$\Lambda_t = sigmoid(\frac{1}{N} \sum_{n=1}^{N} r_{t,n}) \qquad (13)$$

Finally, optimization is performed using cross-entropy loss:

$$L_t = -(\sum_{n=1}^{N} \log g(r_{t,n}, \Lambda_t) + \sum_{n=1}^{N} \log(1 - g(r_{t',n}, \Lambda_t))) \qquad (14)$$

Where $t$ and $t'$ represent two different time steps, $g(r_{t,n}, \Lambda_t) = sigmoid(r_{t,n}^{\mathrm{T}} w_3 \Lambda_t)$, $w_3$ is the learnable matrix.

**Model Training**

We feed the output $x_n^{ST} \in X^{ST}$ of the multi-layer spatio-temporal encoder into the prediction layer MLP. The MLP consists of two fully connected layers, which are responsible for mapping the output of the model's hidden layers to the final prediction result, enabling us to achieve traffic speed prediction for the future time slot $t_p$.

$$\hat{y}_{t_p,n} = MLP(x_n^{ST}) \qquad (15)$$

Then, the model is optimized by minimizing the loss function, aiming to find the combination of weights and biases that minimize the loss. This process improves the accuracy of the predictions.

$$L_p = \sum_{n=1}^{N} \left| y_{t_p,n} - \hat{y}_{t_p,n} \right| \qquad (16)$$

Finally, we sum the self-supervised learning loss and the prediction loss to form the overall loss. Train the model using the backpropagation algorithm until the overall loss converges.

$$L = L_p + L_s + L_t \qquad (17)$$

## EXPERIMENTS

In this section, we conducted extensive experiments on two real-world datasets and visualized the results. Compared to seven baseline models, our proposed model demonstrated state-of-the-art performance in long-term predictions. To validate the effectiveness of each module in our proposed model, we further conducted ablation experiments.

## Datasets

In this paper, we conducted experiments on two real-world datasets, PeMS04 and PeMS08. **TABLE 1** provides information about these two datasets, including the number of nodes, the number of time steps, the sampling interval of time, the number of edges, and the sampling range of time.

**TABLE 1 Dataset description**

| Datasets | Nodes | Time steps | Time intervals | Edges | Time range |
|----------|-------|------------|----------------|-------|------------|
| PeMS04 | 307 | 26208 | 5 min | 340 | 1/1/2018-2/28/2018 |
| PeMS08 | 170 | 17856 | 5 min | 295 | 7/1/2017-8/31/2017 |

## Settings

We implemented the T-ST-SSL model on the PyTorch framework. All experiments were conducted on the NVIDIA GeForce RTX 2080. We split the dataset into training sets, validation sets, and test sets in a 7:1:2 ratio. We trained the model by minimizing the sum of the Mean Absolute Error (MAE) and the loss for the spatial-temporal heterogeneity self-supervised task using the Adam optimizer and backpropagation. During training, the model had a batch size of 32, a learning rate of 0.001, and was trained for 100 epochs. We made predictions for future traffic conditions in the next 30, 45, and 60 minutes using the historical data of the previous 12 observations (60 minutes) on both PeMS04 and PeMS08 datasets.

## Evaluation Metrics

Using multiple evaluation metrics can provide a comprehensive and objective assessment, helping to accurately evaluate the performance of the model and provide guidance for improvement and optimization. In this study, we employed four widely used evaluation metrics in traffic prediction: Mean Absolute Error (MAE) (*23*), Mean Absolute Percentage

Error (MAPE) (*23*), Root Mean Square Error (RMSE) (*24*), and Symmetric Mean Absolute Percentage Error (SMAPE) (*24*). The formulas for these metrics are as follows:

$$MAE = \frac{1}{N} \sum_{n=1}^{N} \left| y_{t_p,n} - \hat{y}_{t_p,n} \right| \qquad (18)$$

$$MAPE = \frac{100\%}{N} \sum_{n=1}^{N} \left| \frac{y_{t_p,n} - \hat{y}_{t_p,n}}{y_{t_p,n}} \right| \qquad (19)$$

$$RMSE = \sqrt{\frac{1}{N} \sum_{n=1}^{N} \left( y_{t_p,n} - \hat{y}_{t_p,n} \right)^2} \qquad (20)$$

$$SMAPE = \frac{100\%}{N} \sum_{n=1}^{N} \frac{\left| y_{t_p,n} - \hat{y}_{t_p,n} \right|}{\left( \left| y_{t_p,n} \right| + \left| \hat{y}_{t_p,n} \right| \right) / 2} \qquad (21)$$

Where $y_{t_p,n}$ and $\hat{y}_{t_p,n}$ represent the ground truth and predicted values at road node $n$ in time slot $t_p$, and $N$ represents the total number of road network nodes.

**Performance Comparison of Proposed Model and Baseline Models**

We mainly compared our proposed model against several typical deep learning models, including ST-SSL (*13*), GWNet (*14*), TGCN (*15*), STTN (*18*), STGCN (25) . We also compared against earlier models such as HA and GRU.

**TABLE 2 The performance comparison of different approaches on PeMS04**

| Dataset | Model | 30min | | | | 45min | | | | 60min | | | |
|---------|-------|-------|------|------|-------|-------|------|------|-------|-------|------|------|-------|
| | | MAE | MAPE | RMSE | SMAPE | MAE | MAPE | RMSE | SMAPE | MAE | MAPE | RMSE | SMAPE |
| PeMS04 | HA | 2.398 | 5.117 | 5.432 | 4.627 | 2.821 | 6.217 | 6.375 | 5.446 | 3.019 | 6.691 | 6.776 | 5.815 |
| | GRU | 2.392 | 5.447 | 5.033 | 4.471 | 2.452 | 5.544 | 5.087 | 4.578 | 2.485 | 5.655 | 5.162 | 4.632 |
| | TGCN | 1.956 | 4.055 | 4.416 | 3.701 | 2.306 | 5.042 | 5.204 | 4.389 | 2.581 | 5.904 | 5.761 | 4.925 |
| | STGCN | 2.018 | 4.156 | <u>4.058</u> | 3.877 | 2.454 | 5.523 | 4.836 | 4.575 | 2.704 | 5.579 | 5.141 | 4.999 |
| | Graph WaveNet | 1.918 | 4.276 | 4.431 | 3.618 | 2.069 | 4.687 | 4.759 | 3.909 | 2.132 | 4.904 | 4.928 | 4.036 |
| | STTN | <u>1.782</u> | 3.806 | <u>4.058</u> | <u>3.376</u> | 2.032 | 4.486 | 4.709 | <u>3.838</u> | 2.294 | 5.421 | 5.236 | 4.384 |
| | ST-SSL | 1.813 | <u>3.786</u> | 4.107 | 3.421 | <u>2.026</u> | <u>4.467</u> | <u>4.583</u> | 3.841 | <u>2.063</u> | <u>4.547</u> | <u>4.706</u> | <u>3.896</u> |
| | T-ST-SSL | **1.720** | **3.591** | **3.879** | **3.246** | **1.931** | **4.169** | **4.483** | **3.656** | **2.021** | **4.427** | **4.690** | **3.819** |

**TABLE 3 The performance comparison of different approaches on PeMS08**

| Dataset | Model | 30min | | | | 45min | | | | 60min | | | |
|---------|-------|-------|-------|-------|-------|-------|-------|-------|-------|-------|-------|-------|-------|
| | | MAE | MAPE | RMSE | SMAPE | MAE | MAPE | RMSE | SMAPE | MAE | MAPE | RMSE | SMAPE |
| PeMS08 | HA | 1.816 | 3.577 | 4.161 | 3.278 | 2.031 | 4.051 | 4.643 | 3.677 | 2.231 | 4.457 | 5.044 | 4.036 |
| | GRU | 2.031 | 4.836 | 4.701 | 3.695 | 2.072 | 4.987 | 4.869 | 3.771 | 2.122 | 5.020 | 4.921 | 3.807 |
| | TGCN | 1.572 | 3.217 | 3.609 | 2.891 | 1.841 | 3.782 | 4.215 | 3.378 | 2.044 | 4.332 | 4.642 | 3.756 |
| | STGCN | 1.694 | 3.432 | 3.660 | 3.186 | 1.831 | 3.880 | 3.954 | 3.360 | 1.943 | 4.472 | 4.391 | 3.571 |
| | Graph WaveNet | 1.622 | 3.751 | 4.024 | 3.004 | 1.692 | 4.005 | 4.297 | 3.135 | 1.786 | 4.382 | 4.531 | 3.276 |
| | STTN | 1.482 | 3.181 | 3.547 | 2.746 | 1.692 | 3.735 | 4.091 | 3.121 | 1.820 | 4.131 | 4.429 | 3.322 |
| | ST-SSL | 1.523 | 3.237 | 3.649 | 2.793 | 1.771 | 3.805 | 4.091 | 3.251 | 1.931 | 4.391 | 4.583 | 3.509 |
| | T-ST-SSL | 1.461 | 3.175 | 3.542 | 2.695 | 1.620 | 3.547 | 4.000 | 2.907 | 1.690 | 3.925 | 4.295 | 3.106 |

**TABLE 2** and **TABLE 3** present the comparison results between the T-ST-SSL model and the baseline models on PeMS04 and PeMS08 datasets. In the tables, the bold font indicates the best solution, while the underlined values represent the second-best results. Based on the comparison results from the tables, we can draw the following conclusions:

1) The T-ST-SSL model outperforms the baseline models on all evaluation metrics for both datasets. Compared to the second-best model, the T-ST-SSL model shows an average improvement of 3.55% in MAE and 2.26% in RMSE. This demonstrates the superiority of the T-ST-SSL model in traffic prediction tasks.

2) Deep learning models considering both temporal and spatial features perform better than models considering single features or traditional time series models.

3) As the prediction horizon increases, models incorporating Transformer (such as T-ST-SSL and STTN) perform better. It demonstrates that Transformer possesses strong global modeling capabilities and has an advantage in capturing long-term temporal dependencies.

4) Adding the modeling of spatial and temporal heterogeneity improves the performance of the models, particularly in road networks with complex topology.

In conclusion, the T-ST-SSL model combines the advantages of extracting spatio-temporal features, utilizing Transformer, and modeling spatio-temporal heterogeneity, which results in outstanding performance in long-term traffic speed prediction tasks.

To provide a more intuitive display of the prediction performance of the T-ST-SSL model, we conducted visualization experiments on randomly selected road segments from the PeMS04 and PeMS08 datasets. The traffic speed prediction results of the T-ST-SSL model for 30 minutes, 45 minutes, and 60 minutes were visualized and compared with the predictions of the ST-SSL model. **Figure 3** to **Figure 5** show the visualization of prediction results for three different prediction horizons on PeMS04 dataset, while **Figure 6** to **Figure 8** display the visualization of prediction results for three different prediction horizons on PeMS08 dataset.

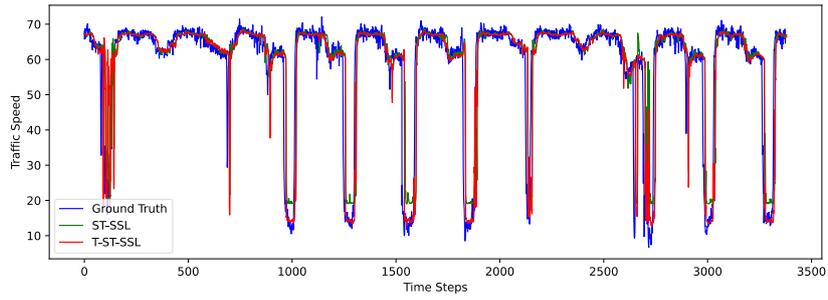

**Figure 3 The visualization results for the prediction horizon of 30 minutes on PeMS04**

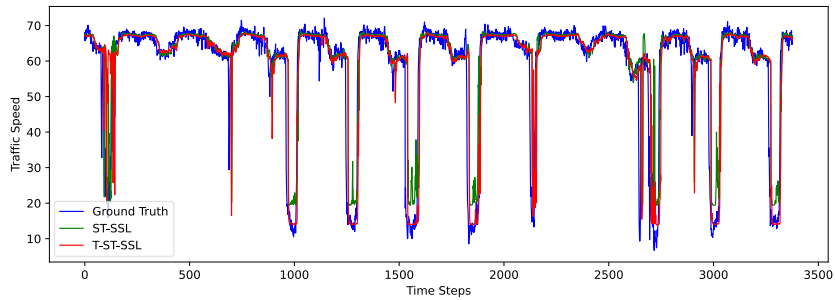

**Figure 4 The visualization results for the prediction horizon of 45 minutes on PeMS04**

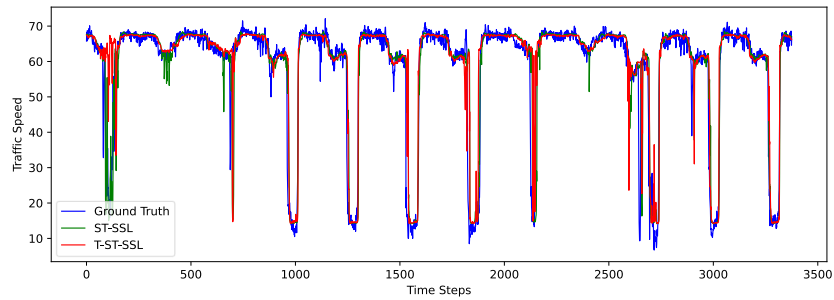

**Figure 5 The visualization results for the prediction horizon of 60 minutes on PeMS04**

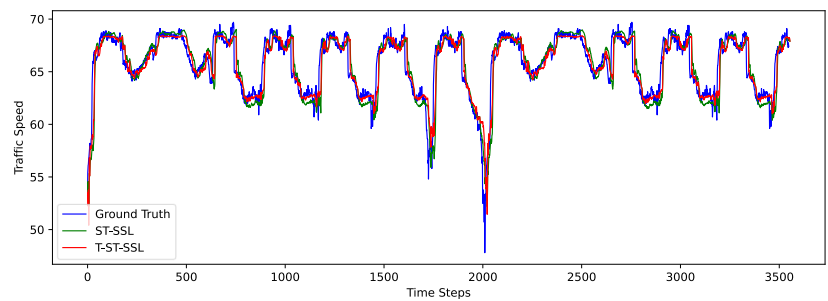

**Figure 6 The visualization results for the prediction horizon of 30 minutes on PeMS08**

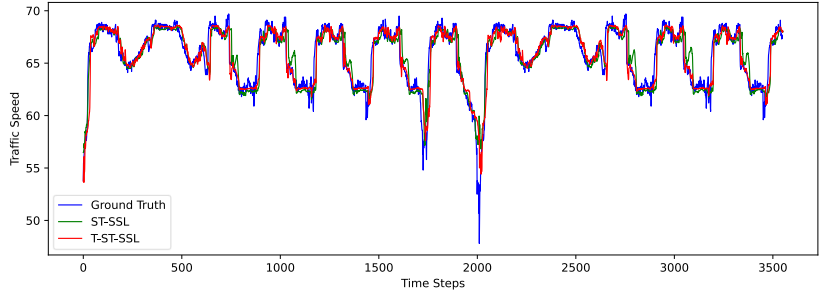

**Figure 7 The visualization results for the prediction horizon of 45 minutes on PeMS08**

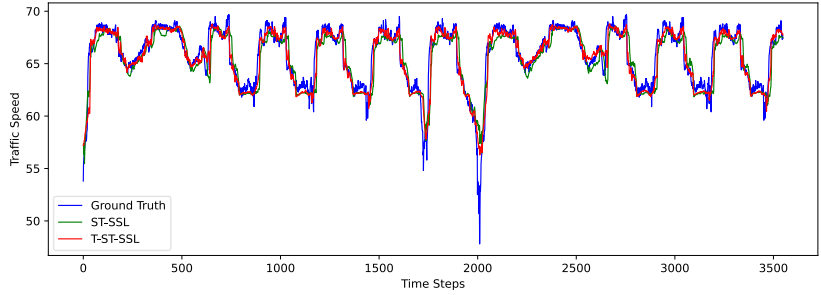

**Figure 8 The visualization results for the prediction horizon of 60 minutes on PeMS08**

In the figures above, we use the blue line to represent the ground truth values of the traffic speed, the green line to represent the predicted values of the ST-SSL model, and the red line to represent the predicted values of the T-ST-SSL model. The results indicate:

1)    In long-term predictions, the T-ST-SSL model consistently outperforms the ST-SSL model, demonstrating that Transformer can effectively capture long-range dependencies by attending to all positions in the sequence simultaneously.

2)    For local extreme points, there is a significant deviation between the predicted values of the ST-SSL model and the ground truth values, especially at local maximum and minimum points. In contrast, the T-ST-SSL model demonstrates superior predictive performance compared to the ST-SSL model, enabling better capture of the changing trends in traffic speed.

3)    In all prediction ranges across the two datasets, the predicted values of the T-ST-SSL model exhibit greater stability and closer proximity to the ground truth values, resulting in better prediction results. This indicates that the T-ST-SSL model is capable of capturing the spatio-temporal dependencies of traffic information more effectively.

**Ablation Experiments**

To further validate the effectiveness of each component of the T-ST-SSL model, we compared it with four variants:

a)    No TH: This variant removes the temporal heterogeneity supervision, meaning that it does not consider the temporal heterogeneity in the traffic sequences.

b)    No SH: This variant removes the spatial heterogeneity supervision, meaning that it does not consider the spatial heterogeneity in the traffic road network.

c)  No T-Block: This variant removes the Transformer in the temporal block, which cannot capture long-term temporal dependencies in the traffic sequences.

d)  No S-Block: This variant removes the graph convolution in the spatial block, which cannot capture long-term dynamic spatial dependencies in the traffic road network.

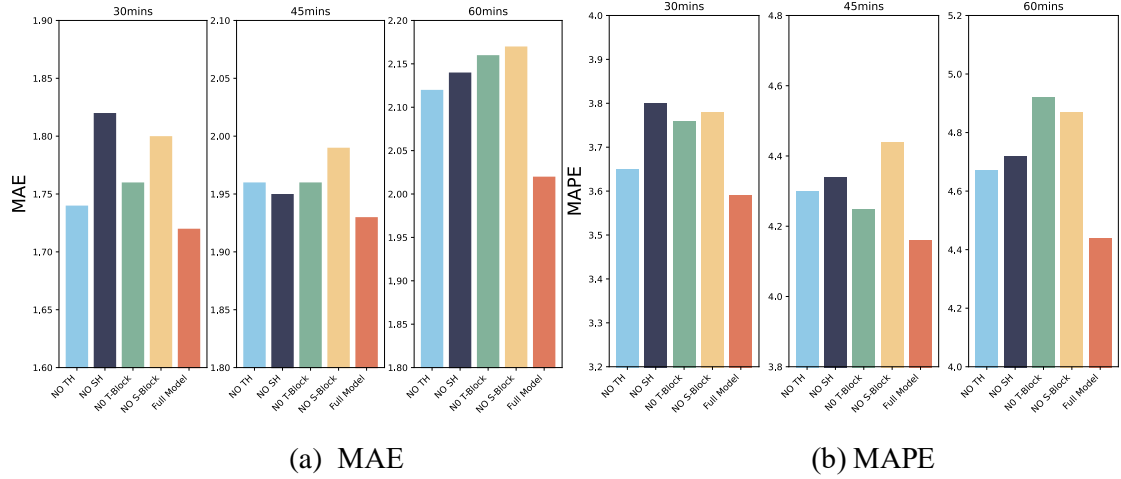

(a)  MAE                                (b)  MAPE

**Figure 9 Prediction metrics of T-ST-SSL and variants on PeMS04 for three horizons**

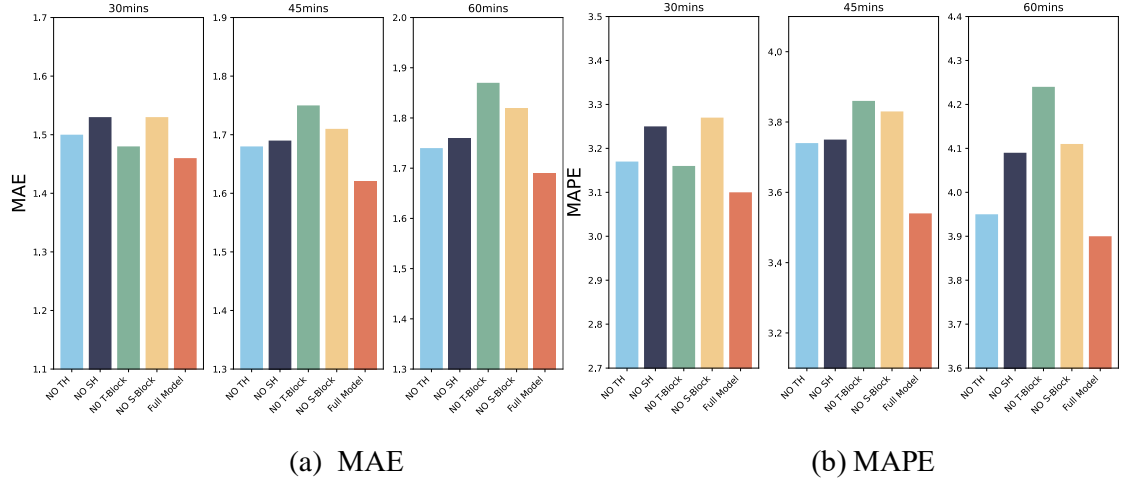

(a)  MAE                                (b)  MAPE

**Figure 10 Prediction metrics of T-ST-SSL and variants on PeMS08 for three horizons**

**Figure 9** and **Figure 10** show the comparison results of these variants on PeMS04 and PeMS08 datasets for three different prediction horizons. The results indicate:

1)  The PeMS04 dataset contains a large number of road network nodes, and the road network's topology is complex. Therefore, the impact of spatial heterogeneity supervision and spatial blocks on the model's performance is more evident. Compared to the T-ST-SSL model, the NO SH and NO S-Block variants have larger prediction errors.

2)  Due to the smaller number of road network nodes in the PeMS08 dataset, it relies more on the model's ability to capture temporal dependencies. Consequently, the variants NO T-Block and NO TH display the larger prediction errors for their lack of temporal blocks and temporal heterogeneity supervision.

3) As the prediction horizon increases on two datasets, the NO T-Block variant shows the largest fluctuation in prediction error and gradually surpasses other variants. This indicates that the Transformer in the temporal block plays a crucial role in capturing long-term temporal dependencies.

In conclusion, each component of the T-ST-SSL model contributes significantly to its performance improvement. The challenge in long-term traffic prediction lies in extracting dynamic long-term temporal and spatial dependencies. By jointly modeling temporal and spatial heterogeneity and incorporating Transformer and graph convolution, the model captures deeper dynamic time and spatial dependencies, thereby enhancing its accuracy.

## CONCLUSIONS

This paper proposes a model that hybrids Transformer and spatio-temporal heterogeneity self-supervised learning for long-term traffic speed prediction. The model utilizes spatio-temporal blocks combining the self-attention mechanism of Transformer and graph convolution to capture the long-term spatio-temporal dependencies in traffic data. To better exploit the spatio-temporal features in the traffic data, we introduce an adaptive data augmentation method that enhances the traffic information at both the sequence level and the graph typological structure level. Additionally, we model the spatio-temporal heterogeneity through self-supervised learning tasks to further enhance the model's performance. Experimental evaluations are conducted on PeMS04 and PeMS08 datasets for three prediction horizons, and the model's performance is evaluated using metrics such as MAE, MAPE, RMSE, and SMAPE. The comparative experiments against seven baseline models consistently demonstrate that the T-ST-SSL model achieves the best prediction performance. Furthermore, we provide a visual analysis of the prediction results, highlighting the robustness of the model and its ability to capture traffic peaks.

In future research, the utilization of multi-source data can be explored to enhance the quality and reliability of traffic data, thereby improving the model's understanding and analysis capabilities for more accurate predictions and better generalization. Additionally, incorporating external environmental information and adaptive adjacency matrix embedding can further enhance the model's ability to capture spatial dependencies.

## ACKNOWLEDGMENTS

This work is supported by China NSFC Program under Grant NO. 61603257.

## AUTHOR CONTRIBUTIONS

The authors' contributions to this paper are as follows: conception and design of the study: W. Z., J. X.; collection of data and implementation of the experiments: W. Z., B. L. D. Z.; analysis and interpretation of results: W. Z.; draft manuscript preparation: W. Z., J. X.; All authors reviewed the results and approved the final version of the paper.